\pdfoutput=1

\documentclass[11pt]{article}

\usepackage{emnlp2021}

\usepackage{times}
\usepackage{latexsym}

\usepackage[T1]{fontenc}

\usepackage[utf8]{inputenc}

\usepackage{microtype}

\usepackage{tcolorbox}
\usepackage[frozencache=true,cachedir=minted-cache]{minted} 
\tcbuselibrary{minted,breakable,xparse,skins,hooks}
\definecolor{bg}{gray}{0.95}
\DeclareTCBListing{mintedbox}{O{}m!O{}}{%
  breakable=true,
  listing engine=minted,
  listing only,
  minted language=#2,
  minted style=default,
  minted options={%
    linenos,
    gobble=0,
    breaklines=true,
    breakafter=,,
    fontsize=\small,
    numbersep=8pt,
    escapeinside=||,
    #1},
  boxsep=0pt,
  left skip=0pt,
  right skip=0pt,
  left=25pt,
  right=0pt,
  top=3pt,
  bottom=3pt,
  arc=5pt,
  leftrule=0pt,
  rightrule=0pt,
  bottomrule=2pt,
  toprule=2pt,
  colback=bg,
  colframe=orange!70,
  enhanced,
  overlay={%
    \begin{tcbclipinterior}
    \fill[orange!20!white] (frame.south west) rectangle ([xshift=20pt]frame.north west);
    \end{tcbclipinterior}},
  #3}

\usepackage{float}
\usepackage{booktabs} 
\usepackage{multirow}
\usepackage{varwidth}
\usepackage{graphicx}
\usepackage{placeins}
\usepackage{colortbl} 
\usepackage{enumitem} 
\usepackage{pifont} 
\usepackage{subcaption} 
\usepackage{cleveref} 

\usepackage{array} 
\newcolumntype{P}[1]{>{\centering\arraybackslash}p{#1}}

\usepackage{stfloats}

\usepackage{pgfplots}
\pgfplotsset{compat=1.17}

\usepackage{etoolbox}
\AtBeginEnvironment{minted}{\singlespacing%
    \fontsize{10}{10}\selectfont}

\title{\textsc{Thermostat}: A Large Collection of \\
NLP Model Explanations and Analysis Tools} 

\author{Nils Feldhus \qquad Robert Schwarzenberg \qquad Sebastian M\"oller \\
  German Research Center for Artificial Intelligence (DFKI) \\
  Speech and Language Technology Lab, Berlin, Germany \\
  \texttt{\{firstname.lastname\}@dfki.de}
  }

\begin{document}
\maketitle

\begin{abstract}

In the language domain, as in other domains, neural explainability takes an ever more important role, with feature attribution methods on the forefront. Many such methods require considerable computational resources and expert knowledge about implementation details and parameter choices. To facilitate research, we present \textsc{Thermostat} which consists of a large collection of model explanations and accompanying analysis tools. \textsc{Thermostat} allows easy access to over 200k explanations for the decisions of prominent state-of-the-art models spanning across different NLP tasks, generated with multiple explainers. The dataset took over 10k GPU hours ($>$ one year) to compile; compute time that the community now saves. The accompanying software tools allow to analyse explanations instance-wise but also accumulatively on corpus level. Users can investigate and compare models, datasets and explainers without the need to orchestrate implementation details. \textsc{Thermostat} is fully open source, democratizes explainability research in the language domain, circumvents redundant computations and increases comparability and replicability. 

\end{abstract}
\begin{figure*}[!ht]
\centering

\begin{subfigure}[t]{0.99\textwidth}

\begin{mintedbox}{python}
import thermostat
data = thermostat.load("imdb-bert-lig")
example = data[0]
\end{mintedbox}

\includegraphics[scale=0.525]{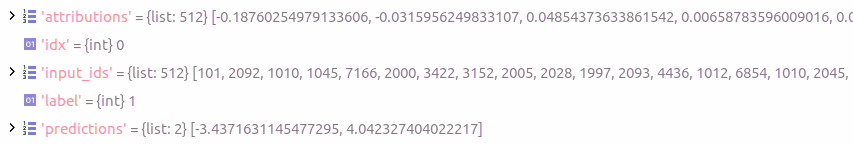}

\end{subfigure}

\vspace*{10pt}
\vspace*{-7.5pt}

\begin{subfigure}[t]{0.99\textwidth}

\begin{mintedbox}{python}
example.render()|\setcounter{FancyVerbLine}{4}|
\end{mintedbox}

\vspace*{5pt}

\includegraphics[scale=0.395]{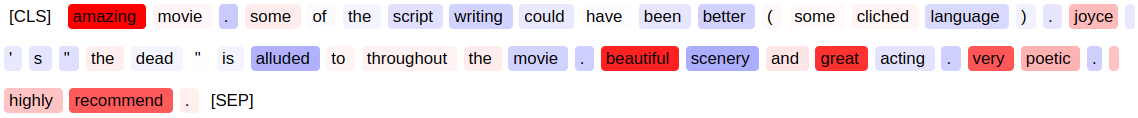}

\end{subfigure}

\caption{Code examples. Top: Loading a dataset and extracting a single instance. Bottom: Visualizing the instance as a heatmap on token level.}
\label{fig:code_load-dataset}

\end{figure*}
\section{Introduction}
\label{sec:introduction}

Deep neural networks are state-of-the-art in natural language processing (NLP) but due to their complexity they are commonly perceived as opaque \cite{Karpathy2015VisualizingAU, Li2016UnderstandingNN}. For this reason, explainability has seen heightened attention in recent years \cite{Belinkov2019AnalysisMI, wallace-etal-2020-interpreting, danilevsky-etal-2020-survey}.

\enlargethispage{\baselineskip}

A prominent class of explainability methods, referred to as \textit{feature attribution methods} (in the following used interchangeably with \textit{explainers}), attributes the output of a complex model to its input features. Feature attribution methods arguably have become a cornerstone of explainability research in NLP: For example, \citet{arras-etal-2016-explaining,arras2017explaining,atanasova-etal-2020-diagnostic,chen-etal-2021-explaining, Neely2021OrderIT} analyze different model architectures with feature attributions. 

There is now also a large body of work comparing explainers in the language domain. 
Explainers are compared with count-based metrics \cite{poerner-etal-2018-evaluating, de-cao-etal-2020-decisions, Tsang2020HowDT, Nguyen2020OnQA, Bodria2021BenchmarkingAS, Ding2021EvaluatingSM, Yin2021OnTF, Hase2021SearchMF, kokhlikyan2021investigating, zafar-etal-2021-lack, sinha2021perturbing} and against human judgement \cite{Nguyen2018ComparingAA, lertvittayakumjorn-toni-2019-human, hase-bansal-2020-evaluating, Prasad2020ToWE}. Feature attribution scores have also been incorporated into model training \cite{Ross2017RightFT, liu-avci-2019-incorporating, Erion2020ImprovingPO, Pruthi2020EvaluatingEH}. 

The feature attribution maps produced and used in the above cited works arguably are the most crucial component of the studies. Unfortunately, none of the above cited papers explicitly links to the generated attribution maps. Easy access to a wide variety of such feature attribution maps, across models, datasets and explainers, however, holds a lot of potential. A central hub
\begin{enumerate}[topsep=0pt,itemsep=-1ex,partopsep=1ex,parsep=1ex]
    \item would increase the comparability and replicability of explainability research, 
    \item would mitigate the computational burden,
    \item would mitigate the implementational burden since in-depth expert knowledge of the explainers and models is required. 
\end{enumerate}
Put differently, a central data hub containing a wide variety of feature attribution maps and offering easy access to them would (1) democratize explainability research to a certain degree, and (2) contribute to green NLP \cite{strubell-etal-2019-energy} and green XAI \cite{Schwarzenberg2021EfficientEF} by circumventing redundant computations. 

\enlargethispage{\baselineskip}

For this reason, we compiled \textsc{Thermostat},\footnote{The term Thermostat is inspired by the Greek word for ``warm'' (``thermos''), hinting at heatmaps being a central application of our contribution. ``stat'' can mean both (1) ``immediately'' referring to the immediate access of attribution maps, and (2) ``statistics'' hinting at cumulative statistics applications.}
an easily accessible data hub that contains a large quantity of feature attribution maps, from numerous explainers, for multiple models, trained on different tasks. Alongside the dataset, we publish a compatible library with analysis and convenience functions. In this paper, we introduce the data hub, showcase the ease of access and discuss use cases.
\section{\textsc{Thermostat}}
\textsc{Thermostat} is intended to be a continuous, collaborative project. As new models, datasets and explainers are published, we invite the community to extend \textsc{Thermostat}. In what follows, we describe the current state which is published under \url{https://github.com/DFKI-NLP/thermostat}.

\label{sec:demonstration}
\subsection{Demonstration}
First, we demonstrate the ease of access. Downloading a dataset requires just two lines of code, as illustrated in the snippet in Fig.~\ref{fig:code_load-dataset}, in which we download the attribution maps as returned by the (Layer) Integrated Gradients explainer \cite{Sundararajan2017AxiomaticAF} for BERT classifications \cite{Devlin2019BERTPO} on the IMDb test set \cite{Maas2011LearningWV}. In addition to the list of feature attributions, the input IDs, the true label of every instance given by the underlying dataset, and the logits of the model's predictions are shipped. Switching the explainer, model or dataset only requires to change the configuration identifier string (``imdb-bert-lig'' in Fig.~\ref{fig:code_load-dataset}). All configuration identifiers in \textsc{Thermostat} consist of three coordinates: dataset, model, and explainer. A visualization tool which returns heatmaps like the one shown in Fig.~\ref{fig:code_load-dataset} is also contained in the accompanying library. 

The object that holds the data after download inherits from the \texttt{Dataset} class of the \texttt{datasets} library \citep{quentin_lhoest_2021_4946100}. This is convenient, because data is cached and versioned automatically in the background and processed efficiently in parallel. Furthermore, \texttt{datasets} provides many convenience functions which can be applied directly, such as filtering and sorting.

\begin{figure*}[t!] 
\centering

\begin{subfigure}[t]{0.99\textwidth}
\centering
\begin{mintedbox}{python}
import thermostat
bert = thermostat.load("multi_nli-bert-occ")
electra = thermostat.load("multi_nli-electra-occ")
disagreement = [(b, e) for (b, e) in zip(bert, electra)
                if b.predicted_label != e.predicted_label]
instance_bert, instance_electra = disagreement[51] # 51: short & interesting
\end{mintedbox}
\caption{Code example that loads two \textsc{Thermostat} datasets. We create a list of instances (\texttt{disagreement}) where the two models (BERT and ELECTRA) do not agree with each other regarding the predicted labels. We then select a demonstrative instance from it.}
\label{fig:code_model-comp}
\end{subfigure}%

\vspace*{5pt}
\begin{subfigure}[t]{0.455\textwidth}
\centering
\begin{mintedbox}{python}
instance_bert.render()|\setcounter{FancyVerbLine}{7}|
\end{mintedbox}
\includegraphics[scale=0.5]{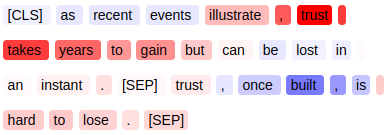}
\caption{Heatmap visualization of the selected instance from Fig.~\ref{fig:code_model-comp}. BERT predicted "entailment" for this example, while the true label is "contradiction".}
\label{fig:mnli-bert}
\end{subfigure}
\hspace*{0.02\textwidth}
\hspace*{0.02\textwidth}
\begin{subfigure}[t]{0.455\textwidth}
\centering
\begin{mintedbox}{python}
instance_electra.render()|\setcounter{FancyVerbLine}{8}|
\end{mintedbox}
\includegraphics[scale=0.5]{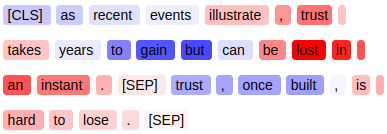}
\caption{Heatmap visualization of the selected instance from Fig.~\ref{fig:code_model-comp}. ELECTRA correctly predicted "contradiction" for this example.}
\label{fig:mnli-electra}
\end{subfigure}


\caption{Code examples for comparing models instance-wise.}
\label{fig:modelcomp}

\end{figure*}

Let us see how this helps us to efficiently compare models and explainers in \textsc{Thermostat}. Let us first compare models. In what follows we consider BERT \cite{Devlin2019BERTPO} and ELECTRA \cite{Clark2020ELECTRAPT}, both trained on the MultiNLI \cite{Williams2018ABC} dataset. We are particularly interested in instances that the two models disagree on. Downloading the explanations and filtering for disagreements is again only a matter of a few lines, as demonstrated in Fig.~\ref{fig:code_model-comp}. 

We derive explanations for the output neuron with the maximum activation. In Fig.~\ref{fig:mnli-bert}, we observe that the Occlusion \cite{Zeiler2014VisualizingAU} explainer does not attribute much importance to the phrase ``can be lost in an instant''. This is plausible since the heatmap explains  a misclassification: the maximum output activation stands for \texttt{entailment}, but the correct label is \texttt{contradiction} and the phrase certainly is a signal for \texttt{contradiction}. In contrast, in the case of ELECTRA (Fig.~\ref{fig:mnli-electra}) which correctly classified the instance the signal phrase receives much higher importance scores. 

After we have now demonstrated how to compare models, let us compare explainers across datasets, as done in previous works. Here, we partly replicate the experiments of \citet{Neely2021OrderIT}. The authors compute the rank correlation in terms of Kendall's $\tau$ between feature attribution maps. If two explainers mostly agree on the importance rank of input features, the value should be high; low otherwise. The authors find that the Integrated Gradients explainer and the LIME \cite{Ribeiro2016WhySI} explainer have a higher $\tau$ value (agree more) for a MultiNLI model (.1794) than when used to rank features for an IMDb-trained classifier (.1050).   

Fig.~\ref{fig:rank-correlation} demonstrates how \textsc{Thermostat} allows to conduct such experiments concisely. The output of the experiment in Fig.~\ref{fig:rank-correlation} reproduces the findings of \citet{Neely2021OrderIT} to a reasonable degree,~i.e.~the $\tau$ value for MultiNLI (.1033) is larger than the $\tau$ value for IMDb (.0257).\footnote{\citet{Neely2021OrderIT} compare DistilBERT \cite{Sanh2019DistilBERTAD} based models, we compare BERT-based models. They further constrain their evaluation to 500 instances while we are calculating the values for the entire datasets.}

\subsection{Maintenance}
\label{sec:easeofuse}

However, explainers such as LIME involve several hyperparameters (number of samples, sampling method, similarity kernel, ...) and thus results can deviate for other choices. \textsc{Thermostat} datasets are versioned and for each version a configuration file is checked in that contains the hyperparameter choices.

If new best practices or bugs emerge, e.g.~in terms of hyperparameters, an updated dataset can be uploaded in a new version. This increases comparability and replicability.

There is also a seamless integration with Hugging Face's \texttt{datasets} as mentioned above. This is why explanation datasets that are published through \texttt{datasets} can be used in \textsc{Thermostat} directly. When contributing new explanation datasets, users simply add the metadata about the three coordinates and make sure that the fields listed in Fig.~\ref{fig:code_load-dataset} are contained. As soon as the dataset is published on the community hub, it can be downloaded and parsed by \textsc{Thermostat}. More details are provided in the repository.\footnote{Users contributing to \textsc{Thermostat} should be aware that the \textsc{Thermostat} project follows the Code of Ethics of ACL and ACM.}
\begin{figure*}[t!] 
\centering

\begin{subfigure}[t]{0.99\textwidth}
\centering
\begin{mintedbox}{python}
import thermostat
from scipy.stats import kendalltau
lime, ig = thermostat.load("imdb-bert-lime"), thermostat.load("imdb-bert-lig")
lime, ig = lime.attributions.flatten(), ig.attributions.flatten()
print(kendalltau(lime, ig))
# KendalltauResult(correlation=0.025657302000906455, pvalue=0.0)
\end{mintedbox}
    
\end{subfigure}%
~
\begin{subfigure}[t]{0.1\textwidth}
\end{subfigure}

\caption{Code example for investigating the rank correlation between LIME and (Layer) Integrated Gradients explanations on IMDb + BERT. The analogous calculation of Kendall's $\tau$ for MultiNLI is left out for brevity. We simply state the value in the last line.}
\label{fig:rank-correlation}

\end{figure*}
\subsection{Current State}

After discussing use and maintenance, we will now present the current state of \textsc{Thermostat}. Please recall that \textsc{Thermostat} is intended to be a continuous, collaborative project.

With the publication of this paper, we contribute the explanations listed in Tab.~\ref{tab:datasetsmodels}. In total, the compute time for generating the explanations already amounts to more than 10,000 GPU hours; computational resources that the community does not have to invest repeatedly now.\footnote{To produce the feature attribution maps, we used up to 24 NVIDIA GPUs in parallel, namely GTX 1080Ti, RTX 2080Ti, RTX 3090, Quadro RTX 6000 and RTX A6000.} Please note that the table is organized around the three coordinates: datasets, models, and explainers, the choices of which we discuss in the following.

\paragraph{Datasets} Currently, four datasets are included in \textsc{Thermostat}, namely IMDb \cite[sentiment analysis,][]{Maas2011LearningWV}, MultiNLI \cite[natural language inference,][]{Williams2018ABC}, XNLI \cite[natural language inference,][]{Conneau2018XNLIEC} and AG News \cite[topic classification,][]{zhang2015character}. We chose these datasets, because arguably, they are prominently used in NLP research.

We hypothesize that instances that the model did not encounter at training time are more informative than known inputs. This is why we concentrated our computational resources on the respective test splits. In total, these amount to almost 50,000 instances already.

\paragraph{Models} The second coordinate in \textsc{Thermostat} is the model. Currently, five model architectures are included, namely 
ALBERT \cite{Lan2020ALBERTAL}, BERT \cite{Devlin2019BERTPO}, ELECTRA \cite{Clark2020ELECTRAPT}, RoBERTa \cite{Liu2019RoBERTaAR}, and XLNet \cite{Yang2019XLNetGA}. We chose community-trained fine-tuned classifiers as they are aptly available through the \texttt{transformers} library \cite{wolf-etal-2020-transformers}, several of which are provided by TextAttack \cite{morris2020textattack}. The repository that we publish can be used to quickly include new models, if they are provided through the \texttt{transformers} library. 

\paragraph{Explainers} Provided through the Captum library \cite{Kokhlikyan2020CaptumAU}, there are five prominent feature attribution methods included in \textsc{Thermostat}. (Layer) Gradient x Activation \cite{Shrikumar2017LearningIF} is an efficient method without hyperparameters. Integrated Gradients \cite{Sundararajan2017AxiomaticAF}, LIME \cite{Ribeiro2016WhySI}, Occlusion \cite{Zeiler2014VisualizingAU} and Shapley Value Sampling \cite{Castro2009PolynomialCO} can be considered computationally challenging and involve hyperparameters. The choice of parameters in \textsc{Thermostat} follows best practices and, as mentioned above, is well-documented and can be updated and extended.  


\begin{table*}[!htbp]
    \begin{tabular}{p{5.25cm} p{1.5cm} p{1cm} p{0.75cm} P{0.7cm} P{0.7cm} P{0.7cm} P{0.7cm} P{0.7cm} }
    \toprule
    Dataset & Split (Subset) & \# instances & \# classes & \multicolumn{5}{c}{Explainers} \\
    \midrule
    
    \textbf{IMDb} & Test & 25000 & 2 & \small{\textbf{LGxA}} & \small{\textbf{LIG}} & \small{\textbf{LIME}} & \small{\textbf{Occ}} & \small{\textbf{SVS}} \\
    
    \multicolumn{4}{r}{\small{
    \texttt{\href{https://huggingface.co/textattack/albert-base-v2-imdb}{textattack/albert-base-v2-imdb}} (ALBERT)}} &
    \cellcolor[rgb]{.1725,.6353,.3725} & 
    \cellcolor[rgb]{.1725,.6353,.3725} & 
    \cellcolor[rgb]{.1725,.6353,.3725} & 
    \cellcolor[rgb]{.1725,.6353,.3725} & 
    \cellcolor[rgb]{.1725,.6353,.3725} \\
    
    \multicolumn{4}{r}{\small{ \texttt{\href{https://huggingface.co/textattack/bert-base-uncased-imdb}{textattack/bert-base-uncased-imdb}} (BERT)}} & 
    \cellcolor[rgb]{.1725,.6353,.3725} & 
    \cellcolor[rgb]{.1725,.6353,.3725} & 
    \cellcolor[rgb]{.1725,.6353,.3725} & 
    \cellcolor[rgb]{.1725,.6353,.3725} &
    \cellcolor[rgb]{.1725,.6353,.3725} \\
    
    \multicolumn{4}{r}{\small{ \texttt{\href{https://huggingface.co/monologg/electra-small-finetuned-imdb}{monologg/electra-small-finetuned-imdb}} (ELECTRA)}} & 
    \cellcolor[rgb]{.1725,.6353,.3725} & 
    \cellcolor[rgb]{.1725,.6353,.3725} & 
    \cellcolor[rgb]{.1725,.6353,.3725} & 
    \cellcolor[rgb]{.1725,.6353,.3725} &
    \cellcolor[rgb]{.1725,.6353,.3725} \\
    
    \multicolumn{4}{r}{\small{
    \texttt{\href{https://huggingface.co/textattack/roberta-base-imdb}{textattack/roberta-base-imdb}} (RoBERTa)}} & 
    \cellcolor[rgb]{.1725,.6353,.3725} &
    \cellcolor[rgb]{.1725,.6353,.3725} &
    \cellcolor[rgb]{.1725,.6353,.3725} & 
    \cellcolor[rgb]{.1725,.6353,.3725} & 
    \cellcolor[rgb]{.1725,.6353,.3725} \\
    
    \multicolumn{4}{r}{\small{
    \texttt{\href{https://huggingface.co/textattack/xlnet-base-cased-imdb}{textattack/xlnet-base-cased-imdb}} (XLNet) }} & 
    \cellcolor[rgb]{.1725,.6353,.3725} &
    \cellcolor[rgb]{.1725,.6353,.3725} &
    \cellcolor[rgb]{.1725,.6353,.3725} & 
    \cellcolor[rgb]{.1725,.6353,.3725} &
    \cellcolor[rgb]{.1725,.6353,.3725} \\
    
    \midrule
    
    \textbf{MultiNLI} & Validation Matched & 9815 & 3 & \small{\textbf{LGxA}} & \small{\textbf{LIG}} & \small{\textbf{LIME}} & \small{\textbf{Occ}} & \small{\textbf{SVS}} \\
    
    \multicolumn{4}{r}{\small{
    \texttt{\href{https://huggingface.co/prajjwal1/albert-base-v2-mnli}{prajjwal1/albert-base-v2-mnli}} (ALBERT) }} &
    \cellcolor[rgb]{.1725,.6353,.3725} & 
    \cellcolor[rgb]{.1725,.6353,.3725} &
    \cellcolor[rgb]{.1725,.6353,.3725} & 
    \cellcolor[rgb]{.1725,.6353,.3725} & 
    \cellcolor[rgb]{.85,.85,.85} \\
    
    \multicolumn{4}{r}{\small{
    \texttt{\href{https://huggingface.co/textattack/bert-base-uncased-MNLI}{textattack/bert-base-uncased-MNLI}} (BERT) }} & 
    \cellcolor[rgb]{.1725,.6353,.3725} & 
    \cellcolor[rgb]{.1725,.6353,.3725} & 
    \cellcolor[rgb]{.1725,.6353,.3725} & 
    \cellcolor[rgb]{.1725,.6353,.3725} & 
    \cellcolor[rgb]{.1725,.6353,.3725} \\
    
    \multicolumn{4}{r}{\small{
    \texttt{\href{https://huggingface.co/howey/electra-base-mnli}{howey/electra-base-mnli}} (ELECTRA) }} &
    \cellcolor[rgb]{.1725,.6353,.3725} &
    \cellcolor[rgb]{.1725,.6353,.3725} &
    \cellcolor[rgb]{.1725,.6353,.3725} & 
    \cellcolor[rgb]{.1725,.6353,.3725} & 
    \cellcolor[rgb]{.1725,.6353,.3725} \\
    
    \multicolumn{4}{r}{\small{
    \texttt{\href{https://huggingface.co/textattack/roberta-base-MNLI}{textattack/roberta-base-MNLI}} (RoBERTa) }} &
    \cellcolor[rgb]{.1725,.6353,.3725} & 
    \cellcolor[rgb]{.1725,.6353,.3725} & 
    \cellcolor[rgb]{.1725,.6353,.3725} &  
    \cellcolor[rgb]{.1725,.6353,.3725} & 
    \cellcolor[rgb]{.1725,.6353,.3725} \\
    
    \multicolumn{4}{r}{\small{
    \texttt{\href{https://huggingface.co/textattack/xlnet-base-cased-MNLI}{textattack/xlnet-base-cased-MNLI}} (XLNet) }} &
    \cellcolor[rgb]{.1725,.6353,.3725} &
    \cellcolor[rgb]{.1725,.6353,.3725} & 
    \cellcolor[rgb]{.1725,.6353,.3725} & 
    \cellcolor[rgb]{.1725,.6353,.3725} &
    \cellcolor[rgb]{.1725,.6353,.3725} \\
    
    \midrule
    
    \textbf{XNLI} & Test (en) & 5010 & 3 & \small{\textbf{LGxA}} & \small{\textbf{LIG}} & \small{\textbf{LIME}} & \small{\textbf{Occ}} & \small{\textbf{SVS}} \\
    
    \multicolumn{4}{r}{\small{
    \texttt{\href{https://huggingface.co/prajjwal1/albert-base-v2-mnli}{prajjwal1/albert-base-v2-mnli}} (ALBERT) }} &
    \cellcolor[rgb]{.1725,.6353,.3725} &
    \cellcolor[rgb]{.1725,.6353,.3725} &
    \cellcolor[rgb]{.1725,.6353,.3725} &
    \cellcolor[rgb]{.1725,.6353,.3725} & 
    \cellcolor[rgb]{.85,.85,.85} \\
    
    \multicolumn{4}{r}{\small{
    \texttt{\href{https://huggingface.co/textattack/bert-base-uncased-MNLI}{textattack/bert-base-uncased-MNLI}} (BERT) }} & 
    \cellcolor[rgb]{.1725,.6353,.3725} &
    \cellcolor[rgb]{.1725,.6353,.3725} & 
    \cellcolor[rgb]{.1725,.6353,.3725} &  
    \cellcolor[rgb]{.1725,.6353,.3725} & 
    \cellcolor[rgb]{.1725,.6353,.3725} \\
    
    \multicolumn{4}{r}{\small{
    \texttt{\href{https://huggingface.co/howey/electra-base-mnli}{howey/electra-base-mnli}} (ELECTRA) }} &
    \cellcolor[rgb]{.1725,.6353,.3725} &
    \cellcolor[rgb]{.1725,.6353,.3725} & 
    \cellcolor[rgb]{.1725,.6353,.3725} &  
    \cellcolor[rgb]{.1725,.6353,.3725} &
    \cellcolor[rgb]{.1725,.6353,.3725} \\
    
    \multicolumn{4}{r}{\small{
    \texttt{\href{https://huggingface.co/textattack/roberta-base-MNLI}{textattack/roberta-base-MNLI}} (RoBERTa) }} &
    \cellcolor[rgb]{.1725,.6353,.3725} &
    \cellcolor[rgb]{.1725,.6353,.3725} & 
    \cellcolor[rgb]{.1725,.6353,.3725} &  
    \cellcolor[rgb]{.1725,.6353,.3725} &
    \cellcolor[rgb]{.85,.85,.85} \\
    
    \multicolumn{4}{r}{\small{
    \texttt{\href{https://huggingface.co/textattack/xlnet-base-cased-MNLI}{textattack/xlnet-base-cased-MNLI}} (XLNet) }} &
    \cellcolor[rgb]{.1725,.6353,.3725} &
    \cellcolor[rgb]{.1725,.6353,.3725} & 
    \cellcolor[rgb]{.1725,.6353,.3725} & 
    \cellcolor[rgb]{.1725,.6353,.3725} & 
    \cellcolor[rgb]{.85,.85,.85} \\
    
    \midrule

    \textbf{AG News} & Test & 7600 & 4 & \small{\textbf{LGxA}} & \small{\textbf{LIG}} & \small{\textbf{LIME}} & \small{\textbf{Occ}} & \small{\textbf{SVS}} \\
    
    \multicolumn{4}{r}{\small{
    \texttt{\href{https://huggingface.co/textattack/albert-base-v2-ag-news}{textattack/albert-base-v2-ag-news}} (ALBERT) }} &
    \cellcolor[rgb]{.1725,.6353,.3725} &
    \cellcolor[rgb]{.1725,.6353,.3725} & 
    \cellcolor[rgb]{.1725,.6353,.3725} &  
    \cellcolor[rgb]{.1725,.6353,.3725} & 
    \cellcolor[rgb]{.1725,.6353,.3725} \\
    
    \multicolumn{4}{r}{\small{
    \texttt{\href{https://huggingface.co/textattack/bert-base-uncased-ag-news}{textattack/bert-base-uncased-ag-news}} (BERT) }} &
    \cellcolor[rgb]{.1725,.6353,.3725} &
    \cellcolor[rgb]{.1725,.6353,.3725} & 
    \cellcolor[rgb]{.1725,.6353,.3725} &  
    \cellcolor[rgb]{.1725,.6353,.3725} & 
    \cellcolor[rgb]{.1725,.6353,.3725} \\
    
    \multicolumn{4}{r}{\small{ \texttt{\href{https://huggingface.co/textattack/roberta-base-ag-news}{textattack/roberta-base-ag-news}} (RoBERTa) }} &
    \cellcolor[rgb]{.1725,.6353,.3725} &
    \cellcolor[rgb]{.1725,.6353,.3725} & 
    \cellcolor[rgb]{.1725,.6353,.3725} &  
    \cellcolor[rgb]{.1725,.6353,.3725} & 
    \cellcolor[rgb]{.1725,.6353,.3725} \\
    
    \bottomrule
    \end{tabular}
    \caption{Overview of feature attribution maps in \textsc{Thermostat}. Dark green cells (86 out of 90) denote available configurations. Gray cells denote configurations that are work-in-progress.}
    \label{tab:datasetsmodels}
\end{table*}

\section{Related Work} \label{sec:relatedwork}

To the best of our knowledge, the closest work to our contribution is the Language Interpretability Tool (LIT) by \citet{Tenney2020TheLI} which offers a graphical interface for exploring saliency maps, counterfactual generation and the visualization of attention and embeddings. We also draw connections to \citet{Hoover2020exBERTAV} and \citet{Lal2021InterpreTAI} who developed interfaces for analyzing the attention mechanism and embeddings of Transformer architectures. These again are visualization and analysis tools and as such complementary to the collection of explanations that is our primary contribution. Thus, an interface that bridges \textsc{Thermostat} and LIT, for instance, is an interesting future direction. 

There exist libraries, such as Captum \cite{Kokhlikyan2020CaptumAU} or transformers-interpret \cite{Pierse2021}, that facilitate the generation of neural explanations. These libraries do not, however, free the user of the computational burden, nor are they easily accessible to non-technical researchers.

As noted in Section \ref{sec:demonstration}, the \texttt{datasets} library \cite{quentin_lhoest_2021_4946100} functions as the backbone of \textsc{Thermostat}. The novelties our work brings with it are (1) attributions from a variety of models that took over 10,000 GPU hours to compute in total, and (2) the support for explainability-specific visualizations and statistics like heatmap visualization and rank correlation between multiple explainers. 

Finally, tangentially related to our work are datasets that supply explanations on top of texts and labels, usually collected from human annotators. e-SNLI \cite{Camburu2018eSNLINL} probably is the most famous example in this line of work. The reader is referred to \citet{Wiegreffe2021TeachMT} for a concise survey. In contrast to our work, the above mentioned papers present human ground truths instead of machine-generated explanations.  

\section{Conclusion}

We present \textsc{Thermostat}, an easily accessible data hub containing a large collection of NLP model explanations from prominent and mostly expensive explainers. We demonstrate the ease of access, extensibility and maintainability. New datasets can be added easily. Furthermore, we showcase an accompanying library and outline use cases. Users can compare models and explainers across a variety of datasets. 

\textsc{Thermostat} democratizes explainability research to a certain degree as it mitigates the computational (environmentally and financially) and implementational burden. Machine-generated explanations become accessible to non-technical researchers. Furthermore, comparability and replicability are increased. 

It becomes apparent when consulting the literature in Section \ref{sec:introduction} that interpretation beyond classification (e.g. machine translation) is still an open problem \cite{wallace-etal-2020-interpreting}. Hence, we focus on these four text classification problems that are well-trodden paths.

\section*{Acknowledgements}

We would like to thank Lisa Raithel, Steffen Castle, and David Harbecke for their valuable feedback. This work has been supported by the German Federal Ministry of Education and Research as part of the projects XAINES (01IW20005) and CORA4NLP (01IW20010).

\bibliography{custom}
\bibliographystyle{acl_natbib}

\end{document}